\documentclass[twocolumn]{article}
\usepackage[utf8]{inputenc}

\usepackage{graphicx}
\usepackage{hyperref}
\hypersetup{hidelinks}
\usepackage{subcaption}
\usepackage{xcolor}
\usepackage{amssymb}
\usepackage{multirow}
\usepackage{siunitx}

\title{The Robotic Vision Scene Understanding Challenge}
\author{David Hall, Ben Talbot, Suman Raj Bista, Haoyang Zhang, Rohan Smith, \\
Feras Dayoub, Niko S\"{u}nderhauf\thanks{This research was conducted by the Australian Research Council Centre of Excellence for Robotic Vision (project number CE140100016), and supported by the QUT Centre for Robotics. Email: \href{mailto:d20.hall@qut.edu.au}{\nolinkurl{d20.hall@qut.edu.au}}}}

\renewcommand\footnotemark{}

\begin{document}
\twocolumn
\maketitle

\begin{abstract}
    Being able to explore an environment and understand the location and type of all objects therein is important for indoor robotic platforms that must interact closely with humans.
    However, it is difficult to evaluate progress in this area due to a lack of standardized testing which is limited due to the need for active robot agency and perfect object ground-truth.
    To help provide a standard for testing scene understanding systems, we present a new robot vision scene understanding challenge using simulation to enable repeatable experiments with active robot agency.
    We provide two challenging task types, three difficulty levels, five simulated environments and a new evaluation measure for evaluating 3D cuboid object maps.
    Our aim is to drive state-of-the-art research in scene understanding through enabling evaluation and comparison of active robotic vision systems. 
    % Challenges help drive progress. Nothing exists in this field that does what we want. We created fancy new challenges with simulated environments, two challenge types, three challenge difficulty levels and a brand spanking new evaluation measure. It is awesome and people should totally try it!
\end{abstract}

\section{Introduction}
% \TODO{Basic points to get across:
% \begin{itemize}
%     \item Scene understanding is important for robotics (link to sparse SLAM representations)
%     \item we want to focus on the creation of reliable sparse maps created by active agents ... and frankly no standards currently meet those needs.
%     \item We have created a new challenge to focus on scene understanding for indoor robotics applications
%     \item good things about us:
%     \begin{itemize}
%         \item Simulation enabling repeatable robotics tests with agency
%         \item Two new challenges for scene understanding
%         \item tiered difficulty levels to enable ablation studies and easy entrance levels
%         \item new evaluation measure for semantic SLAM
%     \end{itemize}
% \end{itemize}
% }
% \david{Something seems off with this beginning para. Moving on.}
One of the long-held goals in robotics is to have autonomous systems which can reliably work alongside humans within unstructured environments such as homes and offices.
A basic building block of such a system is scene understanding, or, knowing what objects exist in its environment and where they are.
In research, this problem of scene understanding is typically viewed through the lens of semantic simultaneous localisation and mapping (SLAM) and is tested using standard datasets like KITTI~\cite{Geiger2013IJRR}, Sun RGBD~\cite{song2015sun}, and Scene Net~\cite{McCormac:etal:ICCV2017}.
Whilst these are beneficial for algorithm comparison, they are static datasets showing a fixed sequence of sensor data and miss a critical component of robotic systems, namely active agency.
% Works in this field also focus more on evaluating the accuracy of robot trajectories over the accuracy of the final map\TODO{REF?}.
% Those that do analyse the map, do not require active exploration to find objects and use standard evaluation measures like average precision (AP) which do not enable fine-grained analysis.

\begin{figure}[t]
    \centering
    \includegraphics[width=\linewidth]{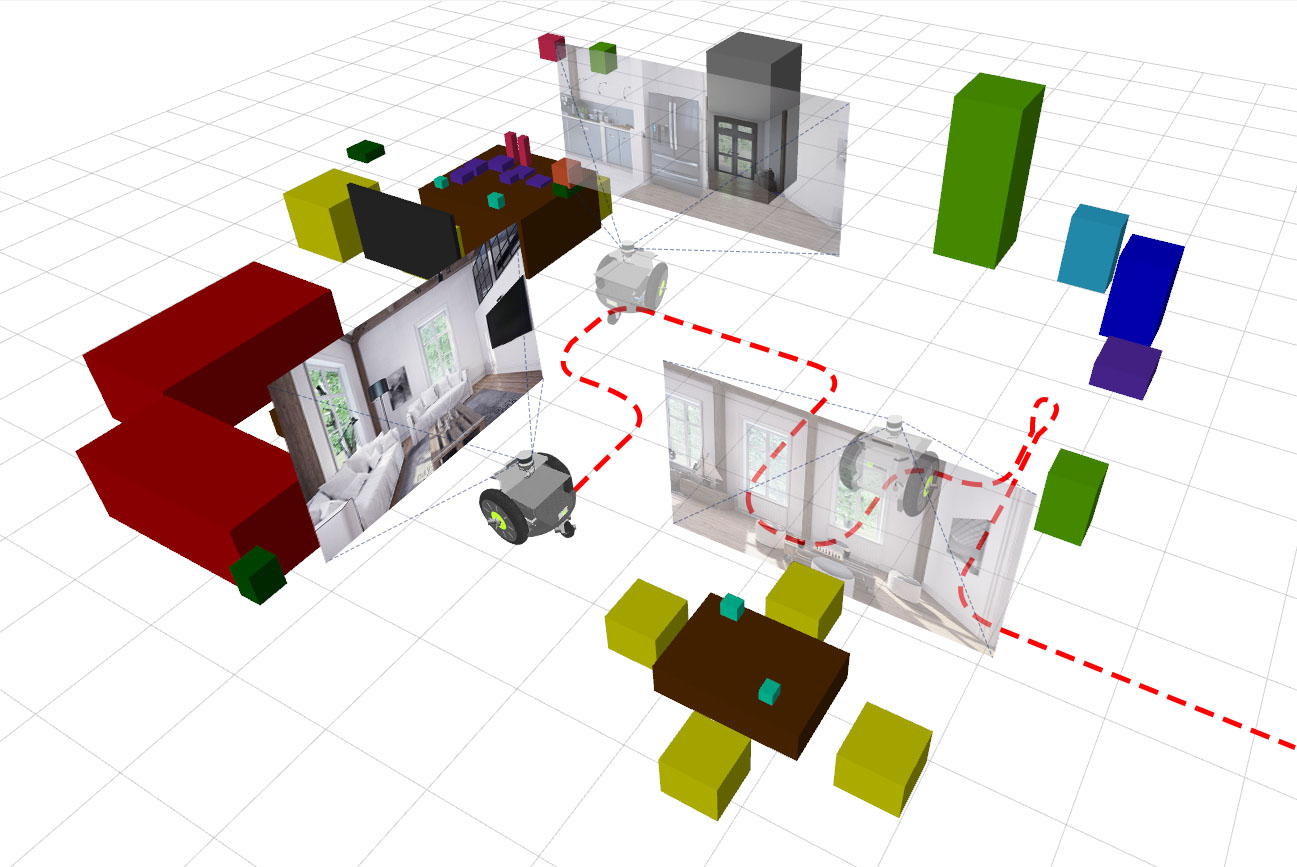}
    \caption{Visual representation of the new scene understanding challenge. A robotic agent must actively explore an environment to produce a 3D cuboid map of all objects of interest. Image by \url{shgraphicdesign.net}}
    \label{fig:su_overview}
\end{figure}

To address this issue, we present a new benchmark challenge for active robotic vision research in scene understanding to enable easy comparison of active systems.
The challenge is summarized visually in Figure~\ref{fig:su_overview}. 
% We utilise simulation to provide consistency, robot agency and accurate ground-truth.
Key aspects and contributions of our new challenge are as follows:
% In our work, we provide a new benchmark challenge for robotics research in semantic object scene understanding, focused on the creation of reliable sparse semantic maps by active robot agents.
% This consists of two challenging tasks: Semantic SLAM and Scene Change Detection (SCD).
% Key contributions of our challenge are as follows:
\begin{itemize}
    \item Definition of two scene understanding tasks: Semantic SLAM and Scene Change Detection (SCD).
    \item Use of simulation to enable repeatable evaluation of active robotic systems
    \item Use of tiered difficulty levels to enable ablation studies and easy particiaption by non-roboticists
    \item Use of discretised action space controlled through a simple API
    \item Use of a new evaluation measure for evaluating object maps from Semantic SLAM
\end{itemize}

% In the remaining sections, we outline: the related work in the field of robotics benchmarking and semantic SLAM evaluation, our specific challenge tasks and difficulty levels, our simulated challenge environment and robot agent setup, and our new evaluation process.

% \david{Need to decide if structure works better with challenge or simulator first. Have paper outline commented out below if desired.}
% Our paper is structured as follows.
% First we outline our two scene understanding challenge tasks and the difficulty levels they can be performed under.
% Second, we outline the simulator used to facilitate the challenges, including details on environment setup, agent control and available sensors.
% Third, we outline our new evaluation measure to be used in comparing algorithm performance at our scene understanding tasks.
% Finally, we summarise our challenge and outline directions for future work in this space.

\section{Related Work}
% \TODO{I think there are two paths we want for the related work. First is how are scene understanding works evaluated/compared now (Semantic SLAM works). The second is the more general, question on how robotic vision challenges more generally are made (can probably nick things from BenchBot paper) Might ask for assistance from Suman and/or Niko}
Our scene understanding challenges intersects two areas of research.
Namely, the creation of standardized challenges and benchmarks for robotics, and the evaluation of sparse semantic SLAM systems.

\subsection{Robotics challenges}
Quantitative evaluation and comparison is uncommon in the field of robotics research when compared to computer vision, due largely to difficulties in standardization of hardware, software, and/or environments~\cite{del2006benchmarks}.
Due to this, robotics challenges typically focus on adding constraints and standards to scenarios that enable comparison.

A common constraint placed on robotics challenges is to use only pre-recorded data.
This enables analysis of an algorithms ability to interpret data for a given task.
This is the approach used by KITTI~\cite{Geiger2013IJRR}, Sun RGBD~\cite{song2015sun}, Cityscape~\cite{Cordts2016Cityscapes}, SceneNet~\cite{McCormac:etal:ICCV2017}, and Oxford RobotCar~\cite{RobotCarDatasetIJRR} datasets which enable multiple tasks to be evaluated such as object tracking, visual odometry, SLAM, semantic segmentation, etc.
However, the constraint of using pre-recorded data is limiting, as a critical component of robotics is that the system is an active agent in its environment, capable of making decisions based upon their observations.

To enable active agents and perform holistic testing of a robot system at a task, some challenges choose to only constrain the testing environment.
Challenges such as RoboCup@Home~\cite{rulebook_2019}, the DARPA challenge~\cite{krotkov2017darpa}, and the Amazon picking challenge~\cite{correll2016analysis} have a fixed environment where competitors can bring their robot systems on a given challenge day for comparison and ranking of a specific problem.
This method provides the most flexibility but challenges are infrequent, have high costs for hadrware, engineering resources and potentially travel, and allow mostly for system-level analysis and comparison rather than any specific area of robotics research.

To analyse specific areas of research in a standardized way with active agency included, some approaches provide access to fixed robot platforms available in a fixed set of environments.
This can be done by remote access to a platform as done by RoboThor~\cite{deitke2020robothor}, iGibson~\cite{xia2020interactive}, and the real robot challenge~\footnote{https://real-robot-challenge.com/en} and is a very promising area of robotics testing research.
However, currently it is more common to use simulation tools to simulate both robots and environments.
This is used not only by systems designed only to work in simulation like AIHabitat~\cite{habitat19iccv}, AI2Thor~\cite{kolve2017ai2}, embodied question answering~\cite{eqa_matterport}, and the active vision dataset~\cite{active-vision-dataset2017}, but also by systems that enable sim-to-real transfer like AirSim~\cite{airsim2017fsr}, CARLA~\cite{Dosovitskiy17}, Isaac~\cite{Isaac}, and the aforementioned, iGibson~\cite{xia2020interactive} and RoboThor~\cite{deitke2020robothor}.
Simulation enables rapid prototyping and comparison of robotic algorithms which consider the active nature of robotics without high implementation costs and, when using the correct tools, without sacrificing the ability to transfer techniques to robot platforms in the real world.
Simulation therefore is the tool we use for our challenge.

\subsection{Semantic SLAM Evaluation}
% While some standardized challenges exist for robotics, the methodologies for evaluating semantic SLAM systems do not focus on the quality of sparse object maps.
% \david{Should maybe be more general about Scene understanding? Thought was to focus on something well known like semantic SLAM, look at how such systems are used and evaluated and why they might not cut it (focus on full point clouds making sparse SLAM representations harder to use, focus on trajectories more than maps).}

Common metrics of evaluating visual SLAM methods are absolute
trajectory error (ATE) and the relative pose error (RPE)~\cite{sturm2012benchmark}.
These are focused on trajectory estimation, comparing global positional offsets, and local motion errors respectively, between estimated and ground-truth trajectories.
These are typically expressed as a root-mean-square-error (RMSE) measure~\cite{sturm2012benchmark,mur2017orb}.
These are used frequently as at least part of a SLAM evaluation process.
While beneficial in testing localisation, these are not required for our purposes as we are focused on the quality of object mapping rather than robot trajectory.

Densely reconstructed SLAM maps are commonly evaluated in a pixel-wise fashion.
When performing a purely spatial analysis, mean distances~\cite{whelan2016elasticfusion} and absolute relative depth distance~\cite{czarnowski2020deepfactors} have been used to evaluate maps.
If semantics are considered, the two major approaches are to either reproject the classified 3D points into 2D and use 2D semantic segmentation evaluation measures, or to adapt 2D measures for a 3D space.
The reprojection approach is seen in~\cite{runz2018maskfusion, hoang2019high} and the adaptation of 2D measures for 3D is shown in~\cite{grinvald2019volumetric, narita2019panopticfusion, hachiuma2019detectfusion}.
These either supplant purely spatial analyses or add to them with extra semantic analysis.
While useful in their given fields, we are focused on sparse object maps that lend themselves easily to rapidly optimized large-scale SLAM systems~\cite{kummerle2general, kaess2011isam2, krauthausen2006exploiting}.

Sparse object-oriented semantic SLAM systems are a relatively recent addition in the field of SLAM research, but contain similar splits in evaluation process as well as utilisng some methods found in object detection research.
If object semantics are not evaluated as part of the map, accuracy of centroid estimation and 3D intersection over union (IoU) are frequently used to evaluate the spatial quality of the map~\cite{geiger2012we, nicholson2018quadricslam, liao2020object}.
In addition, if the representations are not axis aligned, measures such as average orientation similarity~\cite{mousavian20173d,geiger2012we}, and average angular offsets of the main axis~\cite{liao2020object} are additional measures when using bounding box and quadric object representations respectively.
Borrowing from object detection literature, the fusion of semantic and spatial analysis is typically done using adaptations of mean average precision (mAP) with 3D IoU used in place of 2D IoU~\cite{yang2019cubeslam, hou20193d, qi2019deep, yang20203dssd}.
This is effective in combining these two important aspects of sparse object-oriented mapping, however, mAP has some drawbacks that might be detrimental within this setting.
It cannot be broken down to show performance at the disparate aspects of the map (spatial and semantic quality), depends on a tunable threshold to define success, and can obscure false positive object proposals~\cite{hall2020probabilistic}.
This final point can be doubly critical for evaluating object maps, which, when compared to object detection datasets numbering in the many thousands of images and objects, are smaller in scale which has been suggested could increase the magnitude of said problems~\cite{hall2020probabilistic}.
Due to these known weaknesses in mAP, particularly with respect to how they would interplay with the scene understanding challenges we propose, we are led to design our own evaluation measure outlined in Section~\ref{sec:omq}.

\section{Challenges}
In order to address the limited tools available for reliable evaluation of active robotic systems for sparse semantic scene understanding, we present new scene understanding challenges.
We define new semantic SLAM and scene change detection (SCD) scene understanding tasks to help drive robotics research.
We also define three different ``difficulty levels'' to enable participation by non-roboticists, whilst also enabling ablation studies for scene understanding systems.

\begin{figure*}[t]
    \centering
    \begin{subfigure}[b]{0.48\linewidth}
        \centering
        \includegraphics[width=\textwidth]{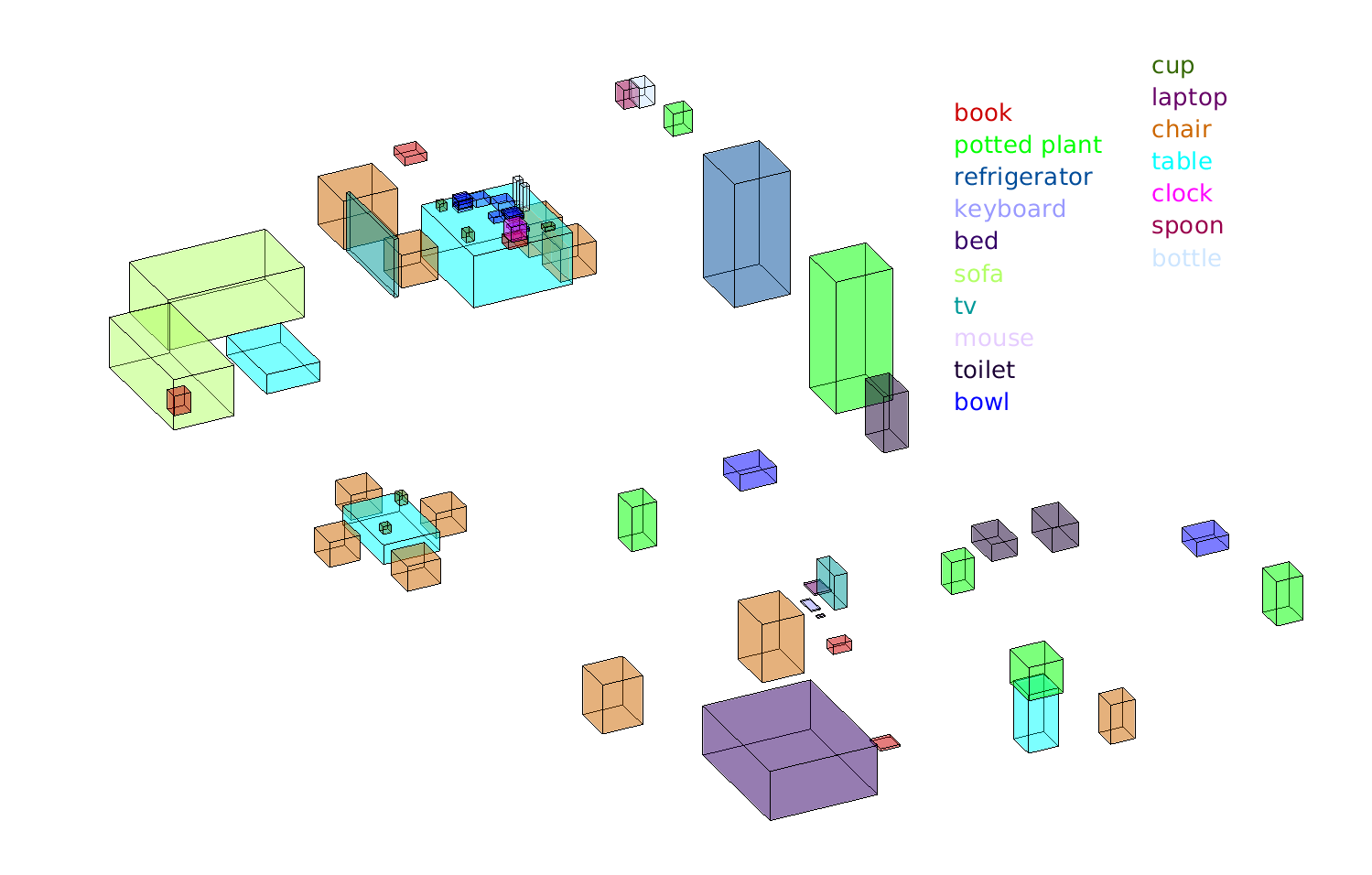}
        \caption{Semantic SLAM output map}
        \label{fig:sem_slam_map}
    \end{subfigure}
    \begin{subfigure}[b]{0.48\linewidth}
        \centering
        \includegraphics[width=\textwidth]{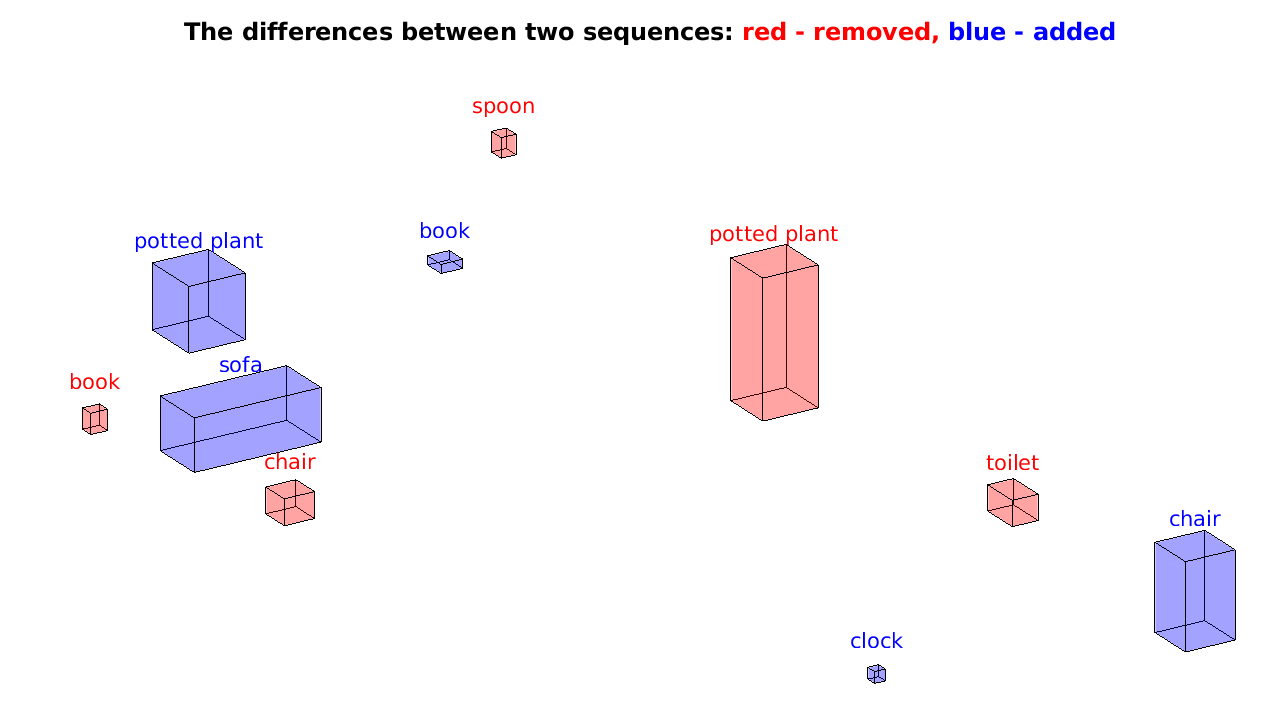}
        \caption{Scene change detection output map}
        \label{fig:scd_map}
    \end{subfigure}
    \caption{Examples of output from the two scene understanding tasks. In semantic SLAM (a) all object axis-aligned cuboids are mapped (colours represent classes). In scene change detection (b) only changed objects are mapped (changes shown by colour) but classes are still required.}
    \label{fig:challenge_maps}
\end{figure*}

\subsection{Semantic SLAM}
% \david{rethink the chosen notation? Is it necessary here?}
The semantic SLAM challenge requires a robotic agent to explore a given environment and create an object map of all objects of interest (a pre-defined set of classes).
In this challenge, an object map ($M$), as proposed by the autonomous system, is made up of a set of $m$ objects ($O$) such that $M = \{ O_1, O_2, O_3, ... O_m \}$.
Every proposed object is defined by an axis-aligned cuboid ($C = \{c_x, c_y, c_z, e_x, e_y, e_z\}$), where $c_x, c_y, c_z$ and $e_x, e_y, e_z$ centroid and extent of the object in the global coordinate system, and a class label probability distribution ($\bold{l}$) across all $n$ object classes of interest, such that $O = \{C, \textbf{l}\}$.
% It should be highlighted again that, unlike other Semantic SLAM problems, the final evaluation is based upon the map produced, rather than robot trajectories.
% As will be highlighted later, this may involve more than just image processing, but rather active exploration and searching in order to be successful.
A visual example of a semantic SLAM object map can be seen in Figure~\ref{fig:sem_slam_map}.
Note that for our purposes, the representation of the ground-truth object map ($\hat{M}$) is likewise made up of a set of $n$ ground-truth objects ($\hat{O}$).
The only difference between ground-truth objects and the proposed objects output by an autonomous system is that the ground-truth class label ($\hat{l}$) is known such that $\hat{O} = \{C,\hat{l}\}$.
% \TODO{Describe the semantic SLAM challenge. Agent explores environment to map out all objects of interest onto cuboid map. Include image of output from Semantic SLAM}

\subsection{Scene Change Detection (SCD)}
In scene change detection (SCD), a robot must explore an environment at two different times, mapping any object changes between them.
In this challenge, object changes consist of objects being added and removed from the environment.
We leave objects moving within the environment as a consideration for future challenges.
Like the semantic SLAM task, the goal in SCD is to create an object map.
However, unlike semantic SLAM the object map should only contain changed objects.
The definition of an object now also includes a state probability ($\textbf{s}$) giving the probability that an object was added, removed, or remained the same.
This changes the object definitions for SCD to $O = \{C, \textbf{l}, \textbf{s}\}$ and $\hat{O} = \{C, \hat{l}, \hat{s}\}$ for proposed and ground-truth objects respectively, where $\hat{s}$ is the true state of an object.
A visual example of an object map for SCD is shown in Figure~\ref{fig:scd_map}.
% \TODO{Describe the scene change detection challenge. Agent explores environment to map out all objects twice and must find the object differences.. Include image of perfect output from scene change detection}

\begin{table*}[t]
\centering
\caption{Robot system sub-tasks that must be considered for the three different challenge difficulty levels.}
\begin{tabular}{l|c|c|c|c|}
\cline{2-5}
\multicolumn{1}{c|}{\textbf{}}              & \textbf{\begin{tabular}[c]{@{}c@{}}Object \\ \\ Detection +\\ Mapping\end{tabular}} 
                                            & \textbf{\begin{tabular}[c]{@{}c@{}}Navigation + \\ \\ Exploration\end{tabular}} 
                                            & \textbf{\begin{tabular}[c]{@{}c@{}}Obstacle \\ \\ Avoidance\end{tabular}} 
                                            & \textbf{Localization} \\ \hline
\multicolumn{1}{|l|}{\textbf{Passive, GT}} & \checkmark &            &            &            \\ \hline
\multicolumn{1}{|l|}{\textbf{Active, GT}}  & \checkmark & \checkmark & \checkmark &            \\ \hline
\multicolumn{1}{|l|}{\textbf{Active, DR}}  & \checkmark & \checkmark & \checkmark & \checkmark \\ \hline
\end{tabular}
\label{tbl:challenge_modes}
\end{table*}

\subsection{Difficulty Levels}
% \david{Idea: maybe add an image showing the different difficulty modes visually?}
One of the novel aspects of our scene understanding challenge is the use of varying difficulty levels that not only lower the bar of entry for non-roboticist researchers, but enable some level of ablation study of the robotic system.
The difficulty levels are applied across both the semantic SLAM and SCD tasks.
We provide 3 levels of difficulty in this challenge, with increases in difficulty corresponding with an increased number of sub-tasks which must be considered, and increased similarity to a robotic system.
This approach allows competitors to perform ablation studies for some sub-tasks, enabling determination of what systems might need to be improved upon.
A summary of the difficulty levels and corresponding robotic sub-tasks to be considered is provided in Table~\ref{tbl:challenge_modes}.

The simplest level of difficulty is a passive mode with ground-truth pose (\textbf{Passive, GT}).
At this level, the robotic agent moves between a set of pre-defined nodes.
This removes the need for active agency, navigation, exploration, and obstacle avoidance.
This is the most akin to other challenges and benchmarks which use pre-recorded data, beneficial as a starting point for those who have no experience controlling active agents.
As ground-truth pose is provided at this level, localization also does not need to be considered.
Although a useful entry point, enabling testing of object detection an mapping, this level is not very representative of robotic systems.

The medium difficulty level is an active mode with ground-truth pose (\textbf{Active, GT}).
This level still provides ground-truth pose data to remove the need for localization, but now gives active control of the robot.
This introduces the elements of needing to actively explore to find all objects of interest, as well as to avoid hitting any obstacles while traversing.
While increasing difficulty, active agency is a critical part of robotic systems that must be considered for any system, making this level more representative than the previous one.
However, agents operating at this difficulty are still working under ideal circumstances, always knowing precisely where they are without the need to self-localize.

The highest difficulty level is the one which most closely resembles a real robot system where the agents have active control but use dead reckoning localization (\textbf{Active, DR}).
Without ground-truth pose information, the agent must self-localize using noisy pose data coming from wheel encoders.
Systems that perform well at this difficulty level are considered the systems with the best chance of having similar performance under real-world conditions.
% \TODO{Describe why we have added ablation studies (lowering bar to entry and enable ablation studies). Outline what is involved in them (different problems in scene understanding with a robotic agent being stripped away one by one)}

% \noindent \textbf{Passive with ground-truth}

% \noindent \textbf{Active with ground-truth}

% \noindent \textbf{Active with dead-reckoning}
\section{The Simulator System}
To provide standardised environments for running repeatable experiments with active robot agents, we utilise a high-fidelity simulation system.
The base simulator tool used in our system is the NVIDIA Isaac Simulator\footnote{\url{https://www.nvidia.com/en-us/deep-learning-ai/industries/robotics/}}, version 2019.2 which provides robot simulation within environments rendered using Unreal Engine\footnote{\url{https://www.unrealengine.com}}.
From this, we define the environments and robotic agents used to complete our scene understanding challenges.
% \TODO{Outline why using simulation is important. How it gives us consistency in testing. Talk about using Isaac as the backend to give us realistic robot control and access to high-fidelity simulated environments through Unreal.}

\subsection{Environments}
\begin{figure*}[t]
    \centering
    \includegraphics[width=\textwidth]{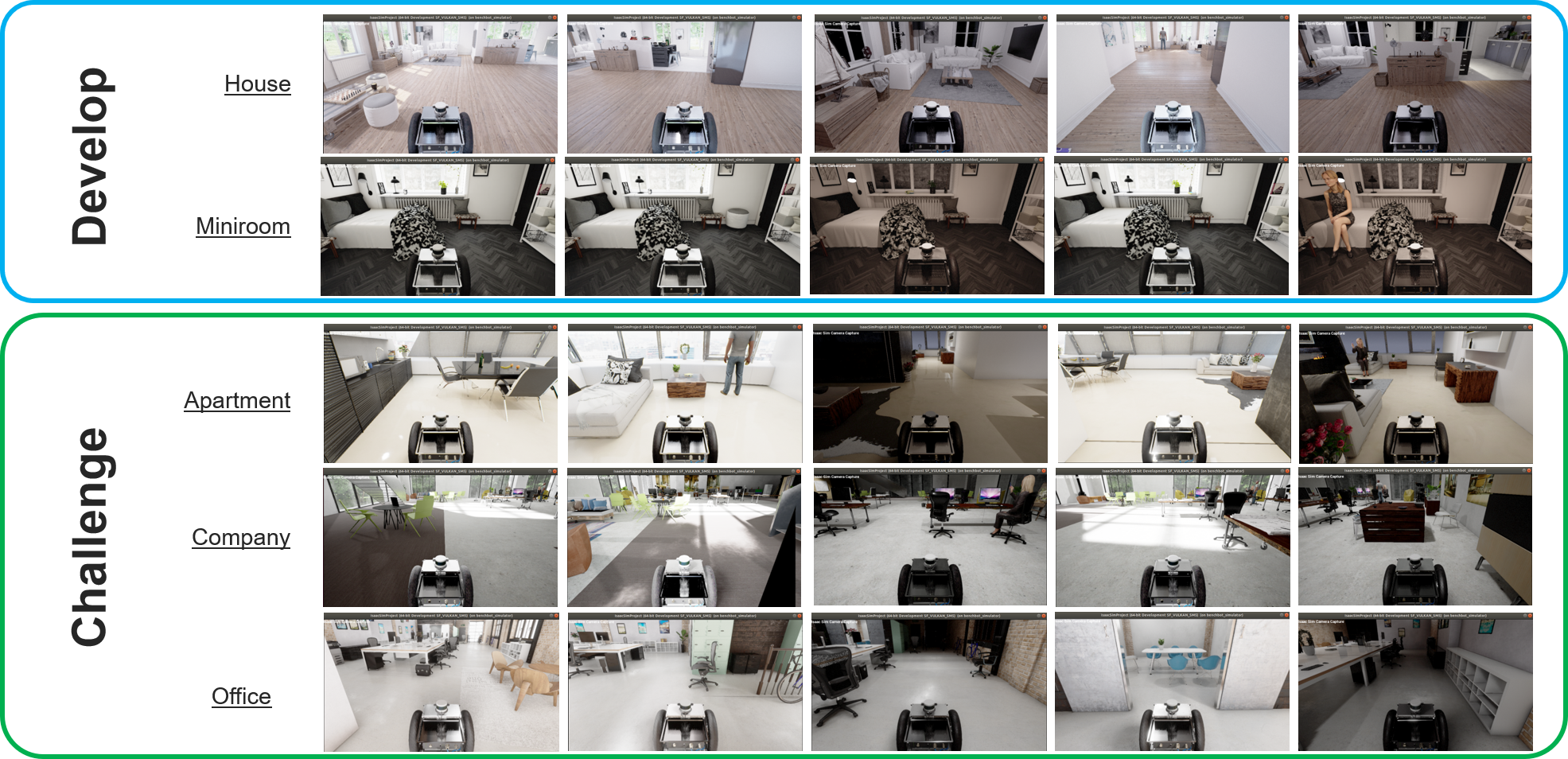}
    \caption{Environments used for scene understanding challenges. There are five variations to each base environment (named). The third and fifth variations show the environments at night.}
    \label{fig:env_splash}
\end{figure*}

Simulated environments consist of five base environments, each having five variations thereof, for a total of 25 environments.
The five base environments are given easily identifiable names: house, miniroom, apartment, company, and office.
The house and miniroom environments are provided for algorithm development, where ground-truth object maps are provided, and apartment, company and office environments are used for final testing.
This provides a total of 10 environments for development and 15 for testing.
It should be noted though that the size of the environments varies, with miniroom being a single small room, whereas house and company are very large environments.
The environments provided are summarized in Figure~\ref{fig:env_splash}.

The base environments are adapted from ones purchased from Evermotion\footnote{\url{https://evermotion.org/}} and adapted to make them suitable for simple exploration by a robotic platform.
The correlation of evermotion environments and their names as given in the challenge are outlined in Table~\ref{tbl:env_names}.
Adaptions made to each base environment to enable exploration include: flattening environments to enable access to elevated areas without stairs, removing doors that block important rooms, and adding walls to section off intraversable areas with built-in objects of interest that would otherwise need to be mapped.
% \david{Worth having before and after images of some of these changes?}

\begin{table}[]
\centering
\caption{Evermotion environments and their corresponding names within the challenge}
\label{tbl:env_names}
\begin{tabular}{|l|l|}
\hline
\multicolumn{1}{|c|}{\textbf{Evermotion}}                             & \textbf{Challenge} \\ \hline
\begin{tabular}[c]{@{}l@{}}Archinteriors\\ Vol 2 Scene 2\end{tabular} & House              \\ \hline
\begin{tabular}[c]{@{}l@{}}Archinteriors\\ Vol 3 Scene 3\end{tabular} & Miniroom           \\ \hline
\begin{tabular}[c]{@{}l@{}}Archinteriors\\ Vol 1 Scene 1\end{tabular} & Apartment          \\ \hline
\begin{tabular}[c]{@{}l@{}}Archinteriors\\ Vol 1 Scene 5\end{tabular} & Company            \\ \hline
\begin{tabular}[c]{@{}l@{}}Archinteriors\\ Vol 1 Scene 3\end{tabular} & Office             \\ \hline
\end{tabular}
\end{table}

Objects of interest are objects that are within a subset of classes from the COCO dataset.
A breakdown of the classes and their frequency across all environment sets (base environment plus variations) is provided in Table~\ref{tbl:obj_dist}.
Objects are either part of these original base environments, or were purchased separately through the Unreal Marketplace.

\begin{table}[]
\centering
\caption{Distribution of all object instances across all environments: Miniroom (M), House (H), Apartment (A), Company (C) and Office (O), and total number of instances across all environments (T). Values for each environment are the total numbers across all 5 variations of each base environment.}
\label{tbl:obj_dist}
\begin{tabular}{|l|l|l|l|l|l|l|}
\hline
\textbf{Class} & \textbf{M} & \textbf{H} & \textbf{A} & \textbf{C} & \textbf{O} & \textbf{T} \\ \hline
\textbf{Bottle}       & 7                 & 15             & 14                 & 1                & 0               & 37             \\ \hline
\textbf{Cup}          & 3                 & 25             & 17                 & 21               & 2               & 68             \\ \hline
\textbf{Bowl}         & 4                 & 42             & 7                  & 12               & 0               & 65             \\ \hline
\textbf{Spoon}        & 3                 & 0              & 0                  & 0                & 0               & 3              \\ \hline
\textbf{Banana}       & 1                 & 0              & 4                  & 0                & 0               & 5              \\ \hline
\textbf{Apple}        & 4                 & 0              & 10                 & 10               & 0               & 24             \\ \hline
\textbf{Orange}       & 3                 & 1              & 1                  & 5                & 3               & 13             \\ \hline
\textbf{Cake}         & 0                 & 0              & 0                  & 3                & 0               & 3              \\ \hline
\textbf{Plant} & 9                 & 32             & 29                 & 47               & 19              & 136            \\ \hline
\textbf{Mouse}        & 0                 & 5              & 0                  & 36               & 40              & 81             \\ \hline
\textbf{Keyboard}     & 0                 & 5              & 0                  & 36               & 40              & 81             \\ \hline
\textbf{Laptop}       & 1                 & 6              & 3                  & 5                & 2               & 17             \\ \hline
\textbf{Book}         & 13                & 25             & 16                 & 18               & 112             & 184            \\ \hline
\textbf{Clock}        & 3                 & 4              & 5                  & 10               & 0               & 22             \\ \hline
\textbf{Chair}        & 10                & 47             & 27                 & 183              & 86              & 353            \\ \hline
\textbf{Table}        & 15                & 21             & 25                 & 57               & 20              & 138            \\ \hline
\textbf{Couch}        & 0                 & 11             & 20                 & 5                & 0               & 36             \\ \hline
\textbf{Bed}          & 5                 & 5              & 0                  & 0                & 0               & 10             \\ \hline
\textbf{Toilet}       & 0                 & 14             & 0                  & 0                & 0               & 14             \\ \hline
\textbf{TV}           & 0                 & 10             & 3                  & 39               & 39              & 91             \\ \hline
\textbf{Microwave}    & 0                 & 0              & 2                  & 0                & 0               & 2              \\ \hline
\textbf{Toaster}      & 2                 & 2              & 4                  & 0                & 0               & 8              \\ \hline
\textbf{Fridge} & 0                 & 3              & 2                  & 4                & 3               & 12             \\ \hline
\textbf{Sink}         & 4                 & 5              & 10                 & 0                & 0               & 19             \\ \hline
\textbf{Person}       & 2                 & 2              & 6                  & 11               & 0               & 21             \\ \hline
\end{tabular}
\end{table}

As previously mentioned, each base environment has a set of five variations.
Each variation has some different combination of objects with between 8 and 27 total object changes (added and removed) between any given pair of environment variations.
As well as this, the third and fifth variation of each environment has been crafted to be an interpretation of the environment at night rather than the day as is the default.
This can be seen in Figure~\ref{fig:env_splash}.
This is to add an extra element of challenge, particularly in SCD where most objects must be re-identified and mapped despite the lighting changes.

We choose environment subsets for the challenge test data for each difficulty level that maximises variation within the subset, and minimises overlap between subsets
The subsets are summarised in Table~\ref{tbl:env_dist}.
Development environments do not have subsets as the intention is for people to experiment as much as they wish with environment combinations when developing solutions and to enable ablation studies using the same environment at different difficulty levels.

\begin{table*}[t]
\caption{Distribution of simulated test environments across task and difficulty level}
\label{tbl:env_dist}
\begin{tabular}{|l|l|l|l|l|l|l|l|l|l|l|l|l|l|l|l|}
\hline
\multirow{2}{*}{\textbf{Task:Difficulty}} & \multicolumn{5}{c|}{\textbf{Office}} & \multicolumn{5}{c|}{\textbf{Apartment}} & \multicolumn{5}{c|}{\textbf{Company}} \\ \cline{2-16} 
  & 1 & 2 & 3 & 4 & 5 & 1 & 2 & 3 & 4 & 5 & 1 & 2 & 3 & 4 & 5     \\ \hline
\textbf{Sem. SLAM: Passive, GT} 
  & \checkmark &            & \checkmark &            &           
  & \checkmark &            & \checkmark &            &            
  & \checkmark &            & \checkmark &            &             \\ \hline
\textbf{Sem. SLAM: Active, GT}          
  &            & \checkmark &            & \checkmark &            
  &            & \checkmark &            & \checkmark &            
  &            & \checkmark &            & \checkmark &            \\ \hline
\textbf{Sem. SLAM: Active, DR}            
  &            &            &            & \checkmark & \checkmark             
  &            &            &            & \checkmark & \checkmark 
  &            &            &            & \checkmark & \checkmark \\ \hline
\textbf{SCD: Passive, GT}
  & \checkmark & \checkmark & \checkmark &            &           
  & \checkmark & \checkmark & \checkmark &            &            
  & \checkmark & \checkmark & \checkmark &            &             \\ \hline
\textbf{SCD: Active, GT}
  &            & \checkmark & \checkmark &            & \checkmark            
  &            & \checkmark & \checkmark &            & \checkmark 
  &            & \checkmark & \checkmark &            & \checkmark  \\ \hline
\textbf{SCD: Active, DR}
  & \checkmark &            &            & \checkmark & \checkmark             
  & \checkmark &            &            & \checkmark & \checkmark 
  & \checkmark &            &            & \checkmark & \checkmark \\ \hline
\end{tabular}
\end{table*}
% In order to ensure a fair division of environment variations across the different challenge task and difficulty levels for both development and final testing

% \TODO{Talk about the environments used in the challenge and how they were created. Here we would talk about using Isaac Sim with Unreal support. Describe changes made to environments, creation of lighting variations, flattening environments. Need Haoyang assistance with environment creation. Maybe include the split of environments for official challenge? (Ben's cool table).}

\subsection{Robot Agents}
Our challenge is made possible through the use of robotic agents that can actively explore an environment to produce a semantic map.
To ensure consistency in the challenge and enable the different challenge difficulties outlined previously, we define fixed sets of discretised robot controls as well as a fixed sensor suite. The sensorimotor interaction between the agent controls and sensors described below, and the underlying robot platform, are provided by the BenchBot system~\cite{talbot2020benchbot}.
% \TODO{This section should describe how we controlled the robot itself. This should reference benchbot but not go into too many details. Possibly outline the process for generating paths for passive mode. Maybe talk a little about collision detection. Largely should focus on what controls the robot has access to, and what the data collected looked like. Maybe some specs about the camera, lidar and relative poses of sensors to robot base?}

\begin{figure*}[t]
    \centering
    \includegraphics[width=0.7\linewidth]{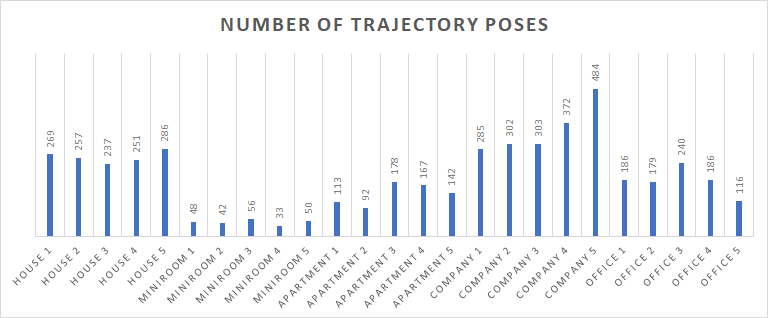}
    \caption{Number of trajectory poses for environments under passive control.}
    \label{fig:pose_chart}
\end{figure*}

\subsubsection{Agent Control}

The challenge provides robot agents with two different types of control mechanisms: passive navigation, and active control. Passive navigation involves guiding a robot through a repeatable, pre-determined trajectory. Active control presents full control of the robot, fulfilling requests to travel explicit linear distances and angular rotations.

Under passive control, the robot travels to the next pose along a trajectory with a guaranteed maximum error of \SI{1}{\cm} in position and \SI{1}{\degree} in orientation. The robot begins inter-pose traversal on a move to next command, employing a pose controller to reconcile differences in current pose and desired pose. Each environment has a different trajectory with length varying from 33 to 484 poses.
The distribution of trajectory lengths can be seen in Figure~\ref{fig:pose_chart}.

Active control allows the robot to respond to navigation requests like ``move forward \SI{1}{\metre}'' and ``rotate \SI{60}{\degree}''. The same controller is employed as in passive control, but a dynamic goal pose is generated given the robot's current position and requested navigation action.

Navigation commands are executed based on the robot's odometry readings which is an important distinction for dead-reckoning modes. For example, in trying to ``move forward \SI{1}{\metre}'' the robot will travel 1m ($\pm\SI{1}{\cm}$) with respect to its odometry readings, but this could be \SI{1.1}{\metre} in reality.

\subsubsection{Agent Sensors}

After each navigation action, the agent is provided with a set of observations from the robot platform's sensor suite. Sensorimotor observations include:
\begin{itemize}
    \item the ground-truth initial pose of the robot in the environment
    \item synced 960x540RGB and depth images from a simulated front-facing sensor
    \item calibration parameters for the RGB and depth sensors, including intrinsic and projection matrices
    \item a \SI{360}{\degree} flat-laser scan from a simulated Velodyne P16 lidar sensor, with a resolution of \SI{0.4}{\degree} and a maximum range of 57.29 m
    \item a list of 6-DOF poses for the platform including the camera, lidar, robot base, and starting pose in the global map frame.
\end{itemize}
\section{Object Map Quality (OMQ) Evaluation}\label{sec:omq}
As our scene understanding challenges focus on the generation of cuboid object maps, we provide a new object map quality (OMQ) evaluation measure to evaluate map accuracy.
The measure is heavily based on the probability-based detection quality (PDQ) evaluation measure designed for probabilistic object detection~\cite{hall2020probabilistic}, now adapted for evaluating 3D object maps for Semantic SLAM.
Adapting OMQ for SCD will be outlined in Section~\ref{sec:omq_scd}.

The process for OMQ is as follows.
For each object in the proposed map, we calculate a pairwise object quality ($pOQ$) between them and each ground-truth object in the ground-truth map. The $pOQ$ score is simply the geometric mean of a spatial and label quality ($Q_{Sp}$ and $Q_L$ respectively). This is formally expressed for the $i$-th proposed object ($O_i$) and $j$-th ground-truth object ($\hat{O}_j$) as

\begin{equation}\label{eqn:poq}
    pOQ(O_i, \hat{O}_j) = \sqrt{Q_{Sp}(O_i, \hat{O}_j) \cdot Q_L(O_i, \hat{O}_j)}.
\end{equation}

Unlike in PDQ, for OMQ spatial quality is simply the 3D IoU score between ground-truth and proposed object cuboids as is used in other works~\cite{geiger2012we, liao2020object}.
Label quallity is calculated in the same manner as PDQ, taking the probability given to the correct class.

\begin{equation}
    Q_L(O_i, \hat{O}_j) = \textbf{l}_i(\hat{l}_j)
\end{equation}

Once all pairwise scores are calculated, objects in the proposed map are optimally assigned to objects in the ground-truth map.
From this we atain a list of ``true positive'' quality scores with non-zero quality assignments ($\textbf{q}^{TP}$), and the number of ``true positives'' ($N_{TP}$), false negatives ($N_{FN}$), and false positives $N_{FP}$.
Unlike in the original PDQ paper~\cite{hall2020probabilistic}, we also create a list of false positive costs ($\textbf{c}_{FP}$) for all false positives in order to weigh overconfident false positives as worse than low-confidence false positives.
This is in alignment with later versions of PDQ found in the latest versions of their code\footnote{\url{https://github.com/jskinn/rvchallenge-evaluation}} but not formally expressed in any paper.
False positive cost is simply the maximum label probability given to a non-background (BG) class label, expressed formally as

\begin{equation}
    FP_{cost}(O_i) = \max_{l \neq \textrm{BG}} \textbf{l}_i(l).
\end{equation}

\noindent This is what is currently used for PDQ in the probabilistic object detection (PrOD) challenges\footnote{\url{https://competitions.codalab.org/competitions/2059}}. 

The final OMQ score is calculated as an average of ``true positive'' pOQ scores, across all ``true positives'' false negatives and false positives, with false positives weighted according to their aforementioned cost.
This is formally expressed as

\begin{equation}
    OMQ(M, \hat{M}) = \frac{\sum_{i=1}^{N_{TP}} \textbf{q}(i)}{N_{TP} + N_{FN} + \sum_{j=1}^{N_{FP}} \textbf{c}_{FP}(j)} .
\end{equation}

\noindent It should be noted that as with PDQ, the spatial and label qualities can be used as separate quality measures for fine-grained analysis.

\subsection{Adapting OMQ for SCD}\label{sec:omq_scd}
While the previous definition of OMQ is suited for Semantic SLAm analysis, the OMQ measure is easily adapted for SCD by adding a state quality term to accommodate the confidence given that an object has been added or removed.
State quality ($Q_{St}$) operates the same way as label quality, simply being the confidence given to the correct state of the object in question (added or removed).This is formally expressed as

\begin{equation}
    Q_{St}(O_i, \hat{O}_j) = \textbf{s}_i(\hat{s}_j).
\end{equation}

The pOQ score for SCD then becomes the geometric mean between the label, spatial, and state qualities as follows

\begin{equation}
    pOQ_{SCD} = \sqrt[3]{Q_{Sp} \cdot Q_L \cdot Q_{St}}.
\end{equation}.

Note that above, all components are equations dependant on $i$-th proposed object and $j$-th ground-truth object as in equation~(\ref{eqn:poq}) but is abbreviated here for spacing reasons.
Also note that including state quality will change the relative importance of the spatial and semantic aspects of quality from that seen in Semantic SLAM due to averaging over three terms instead of two.

State probability also effects the false positive cost of OMQ.
False positive cost becomes the geometric mean of the maximum non-background label probability, and the maximum state probability that is not ``same'' (added or removed).
Formally this is

\begin{equation}
    FP_{SCDcost}(O_i) = \sqrt{\max_{l \neq \textrm{BG}} \textbf{l}_i(l) \cdot \max_{s \neq \textrm{same}} \textbf{s}_i(s)}.
\end{equation}

\noindent With this consideration, the worst false positives are certain about what the object is and that said object was added or removed.
As we use the geometric mean, if either of these is very low, the cost of the false positive will be greatly reduced.

Beyond these changes to false pOMQ and false positive cost, OMQ is calculated as normal for SCD as for Semantic SLAM.
% \TODO{This section should talk all about the OMQ evaluation process. It is an extension to PDQ designed for 3D cuboid evaluation. Use an algorithm flow like was done with PDQ. Explain the ability to partition the measure into multiple parts for more analysis (remember you may be talking to roboticists rather than computer vision researchers). Outline differences between OMQ for semantic SLAM and OMQ for SCD.}
\section{Summary and Future Work}
% \TODO{remind everyone what we have made, what it is for and why it is important. Leave open to future work expanding the dataset of environments and increasing fidelity, enabling robot interaction with objects like in iGibson, and updating evaluation measures to account for localization errors in map making?}
We show a new robotic vision scene understanding challenge utilising high-fidelity simulation to enable active robot agency.
Unlike static datasets which show a pre-recorded set of data, in our challenge the robot agents have autonomy and must actively explore environments to map out objects and detect changes in scenes.
We outline two new challenge tasks, three difficulty levels that enable ablation studies and contributions from non-roboticists, our simulated environments and robtoic agents, and our new object map quality evaluation measure.
Future works should build on the basic ideas covered in this paper, expanding both the realism and variety of environments and robotic platforms used as the current data pool is limited when compared to the variety of static datasets.
Further expansions could utilise ideas such as object interactivity and dynamic obstacles as used in other works.
Finally, providing sim-to-real transfer of solutions, as is made possible by tools like BenchBot, should be persued further to reliably test solutions in real-world settings.

\bibliography{refs}

% Generated by IEEEtran.bst, version: 1.14 (2015/08/26)
\begin{thebibliography}{10}
\providecommand{\url}[1]{#1}
\csname url@samestyle\endcsname
\providecommand{\newblock}{\relax}
\providecommand{\bibinfo}[2]{#2}
\providecommand{\BIBentrySTDinterwordspacing}{\spaceskip=0pt\relax}
\providecommand{\BIBentryALTinterwordstretchfactor}{4}
\providecommand{\BIBentryALTinterwordspacing}{\spaceskip=\fontdimen2\font plus
\BIBentryALTinterwordstretchfactor\fontdimen3\font minus
  \fontdimen4\font\relax}
\providecommand{\BIBforeignlanguage}[2]{{%
\expandafter\ifx\csname l@#1\endcsname\relax
\typeout{** WARNING: IEEEtran.bst: No hyphenation pattern has been}%
\typeout{** loaded for the language `#1'. Using the pattern for}%
\typeout{** the default language instead.}%
\else
\language=\csname l@#1\endcsname
\fi
#2}}
\providecommand{\BIBdecl}{\relax}
\BIBdecl

\bibitem{Geiger2013IJRR}
A.~Geiger, P.~Lenz, C.~Stiller, and R.~Urtasun, ``Vision meets robotics: The
  kitti dataset,'' \emph{International Journal of Robotics Research (IJRR)},
  2013.

\bibitem{song2015sun}
S.~Song, S.~P. Lichtenberg, and J.~Xiao, ``Sun rgb-d: A rgb-d scene
  understanding benchmark suite,'' in \emph{Proceedings of the IEEE conference
  on computer vision and pattern recognition}, 2015, pp. 567--576.

\bibitem{McCormac:etal:ICCV2017}
J.~McCormac, A.~Handa, S.~Leutenegger, and A.~J.Davison, ``Scenenet rgb-d: Can
  5m synthetic images beat generic imagenet pre-training on indoor
  segmentation?'' 2017.

\bibitem{del2006benchmarks}
A.~P. del Pobil, R.~Madhavan, and E.~Messina, ``Benchmarks in robotics
  research,'' in \emph{Workshop IROS}.\hskip 1em plus 0.5em minus 0.4em\relax
  Citeseer, 2006.

\bibitem{Cordts2016Cityscapes}
M.~Cordts, M.~Omran, S.~Ramos, T.~Rehfeld, M.~Enzweiler, R.~Benenson,
  U.~Franke, S.~Roth, and B.~Schiele, ``The cityscapes dataset for semantic
  urban scene understanding,'' in \emph{Proc. of the IEEE Conference on
  Computer Vision and Pattern Recognition (CVPR)}, 2016.

\bibitem{RobotCarDatasetIJRR}
\BIBentryALTinterwordspacing
W.~Maddern, G.~Pascoe, C.~Linegar, and P.~Newman, ``{1 Year, 1000km: The Oxford
  RobotCar Dataset},'' \emph{The International Journal of Robotics Research
  (IJRR)}, vol.~36, no.~1, pp. 3--15, 2017. [Online]. Available:
  \url{http://dx.doi.org/10.1177/0278364916679498}
\BIBentrySTDinterwordspacing

\bibitem{rulebook_2019}
``Robocup@home 2019: Rules and regulations (draft),''
  \url{http://www.robocupathome.org/rules/2019_rulebook.pdf}, 2019.

\bibitem{krotkov2017darpa}
E.~Krotkov, D.~Hackett, L.~Jackel, M.~Perschbacher, J.~Pippine, J.~Strauss,
  G.~Pratt, and C.~Orlowski, ``The darpa robotics challenge finals: Results and
  perspectives,'' \emph{Journal of Field Robotics}, vol.~34, no.~2, pp.
  229--240, 2017.

\bibitem{correll2016analysis}
N.~Correll, K.~E. Bekris, D.~Berenson, O.~Brock, A.~Causo, K.~Hauser, K.~Okada,
  A.~Rodriguez, J.~M. Romano, and P.~R. Wurman, ``Analysis and observations
  from the first amazon picking challenge,'' \emph{IEEE Transactions on
  Automation Science and Engineering}, vol.~15, no.~1, pp. 172--188, 2016.

\bibitem{deitke2020robothor}
M.~Deitke, W.~Han, A.~Herrasti, A.~Kembhavi, E.~Kolve, R.~Mottaghi,
  J.~Salvador, D.~Schwenk, E.~VanderBilt, M.~Wallingford \emph{et~al.},
  ``Robothor: An open simulation-to-real embodied ai platform,'' in
  \emph{Proceedings of the IEEE/CVF Conference on Computer Vision and Pattern
  Recognition}, 2020, pp. 3164--3174.

\bibitem{xia2020interactive}
F.~Xia, W.~B. Shen, C.~Li, P.~Kasimbeg, M.~E. Tchapmi, A.~Toshev,
  R.~Mart{\'\i}n-Mart{\'\i}n, and S.~Savarese, ``Interactive gibson benchmark:
  A benchmark for interactive navigation in cluttered environments,''
  \emph{IEEE Robotics and Automation Letters}, vol.~5, no.~2, pp. 713--720,
  2020.

\bibitem{habitat19iccv}
{Manolis Savva*}, {Abhishek Kadian*}, {Oleksandr Maksymets*}, Y.~Zhao,
  E.~Wijmans, B.~Jain, J.~Straub, J.~Liu, V.~Koltun, J.~Malik, D.~Parikh, and
  D.~Batra, ``Habitat: {A} {P}latform for {E}mbodied {AI} {R}esearch,'' in
  \emph{Proceedings of the IEEE/CVF International Conference on Computer Vision
  (ICCV)}, 2019.

\bibitem{kolve2017ai2}
E.~Kolve, R.~Mottaghi, W.~Han, E.~VanderBilt, L.~Weihs, A.~Herrasti, D.~Gordon,
  Y.~Zhu, A.~Gupta, and A.~Farhadi, ``Ai2-thor: An interactive 3d environment
  for visual ai,'' \emph{arXiv preprint arXiv:1712.05474}, 2017.

\bibitem{eqa_matterport}
E.~Wijmans, S.~Datta, O.~Maksymets, A.~Das, G.~Gkioxari, S.~Lee, I.~Essa,
  D.~Parikh, and D.~Batra, ``{E}mbodied {Q}uestion {A}nswering in
  {P}hotorealistic {E}nvironments with {P}oint {C}loud {P}erception,'' in
  \emph{Proceedings of the IEEE Conference on Computer Vision and Pattern
  Recognition (CVPR)}, 2019.

\bibitem{active-vision-dataset2017}
P.~Ammirato, P.~Poirson, E.~Park, J.~Kosecka, and A.~C. Berg, ``A dataset for
  developing and benchmarking active vision,'' in \emph{IEEE International
  Conference on Robotics and Automation (ICRA)}, 2017.

\bibitem{airsim2017fsr}
\BIBentryALTinterwordspacing
S.~Shah, D.~Dey, C.~Lovett, and A.~Kapoor, ``Airsim: High-fidelity visual and
  physical simulation for autonomous vehicles,'' in \emph{Field and Service
  Robotics}, 2017. [Online]. Available: \url{https://arxiv.org/abs/1705.05065}
\BIBentrySTDinterwordspacing

\bibitem{Dosovitskiy17}
A.~Dosovitskiy, G.~Ros, F.~Codevilla, A.~Lopez, and V.~Koltun, ``{CARLA}: {An}
  open urban driving simulator,'' in \emph{Proceedings of the 1st Annual
  Conference on Robot Learning}, 2017, pp. 1--16.

\bibitem{Isaac}
\BIBentryALTinterwordspacing
{NVIDIA Corporation}, ``Nvidia isaac: The platform for robotics,'' 2019,
  accessed: 31-07-20. [Online]. Available:
  \url{https://www.nvidia.com/en-au/deep-learning-ai/industries/robotics/}
\BIBentrySTDinterwordspacing

\bibitem{sturm2012benchmark}
J.~Sturm, N.~Engelhard, F.~Endres, W.~Burgard, and D.~Cremers, ``A benchmark
  for the evaluation of rgb-d slam systems,'' in \emph{2012 IEEE/RSJ
  International Conference on Intelligent Robots and Systems}.\hskip 1em plus
  0.5em minus 0.4em\relax IEEE, 2012, pp. 573--580.

\bibitem{mur2017orb}
R.~Mur-Artal and J.~D. Tard{\'o}s, ``Orb-slam2: An open-source slam system for
  monocular, stereo, and rgb-d cameras,'' \emph{IEEE Transactions on Robotics},
  vol.~33, no.~5, pp. 1255--1262, 2017.

\bibitem{whelan2016elasticfusion}
T.~Whelan, R.~F. Salas-Moreno, B.~Glocker, A.~J. Davison, and S.~Leutenegger,
  ``Elasticfusion: Real-time dense slam and light source estimation,''
  \emph{The International Journal of Robotics Research}, vol.~35, no.~14, pp.
  1697--1716, 2016.

\bibitem{czarnowski2020deepfactors}
J.~Czarnowski, T.~Laidlow, R.~Clark, and A.~J. Davison, ``Deepfactors:
  Real-time probabilistic dense monocular slam,'' \emph{IEEE Robotics and
  Automation Letters}, vol.~5, no.~2, pp. 721--728, 2020.

\bibitem{runz2018maskfusion}
M.~Runz, M.~Buffier, and L.~Agapito, ``Maskfusion: Real-time recognition,
  tracking and reconstruction of multiple moving objects,'' in \emph{2018 IEEE
  International Symposium on Mixed and Augmented Reality (ISMAR)}.\hskip 1em
  plus 0.5em minus 0.4em\relax IEEE, 2018, pp. 10--20.

\bibitem{hoang2019high}
D.-C. Hoang, T.~Stoyanov, and A.~J. Lilienthal, ``High-quality instance-aware
  semantic 3d map using rgb-d camera,'' \emph{arXiv preprint arXiv:1903.10782},
  2019.

\bibitem{grinvald2019volumetric}
M.~Grinvald, F.~Furrer, T.~Novkovic, J.~J. Chung, C.~Cadena, R.~Siegwart, and
  J.~Nieto, ``Volumetric instance-aware semantic mapping and 3d object
  discovery,'' \emph{IEEE Robotics and Automation Letters}, vol.~4, no.~3, pp.
  3037--3044, 2019.

\bibitem{narita2019panopticfusion}
G.~Narita, T.~Seno, T.~Ishikawa, and Y.~Kaji, ``Panopticfusion: Online
  volumetric semantic mapping at the level of stuff and things,'' \emph{arXiv
  preprint arXiv:1903.01177}, 2019.

\bibitem{hachiuma2019detectfusion}
R.~Hachiuma, C.~Pirchheim, D.~Schmalstieg, and H.~Saito, ``Detectfusion:
  Detecting and segmenting both known and unknown dynamic objects in real-time
  slam,'' \emph{arXiv preprint arXiv:1907.09127}, 2019.

\bibitem{kummerle2general}
R.~K{\"u}mmerle, G.~Grisetti, H.~Strasdat, K.~Konolige, and W.~Burgard, ``o: A
  general framework for graph optimization,'' in \emph{Proceedings of the 2011
  IEEE International Conference on Robotics and Automation}, 2, pp. 3607--3613.

\bibitem{kaess2011isam2}
M.~Kaess, H.~Johannsson, R.~Roberts, V.~Ila, J.~Leonard, and F.~Dellaert,
  ``isam2: Incremental smoothing and mapping with fluid relinearization and
  incremental variable reordering,'' in \emph{2011 IEEE International
  Conference on Robotics and Automation}.\hskip 1em plus 0.5em minus
  0.4em\relax IEEE, 2011, pp. 3281--3288.

\bibitem{krauthausen2006exploiting}
P.~Krauthausen, A.~Kipp, and F.~Dellaert, ``Exploiting locality in slam by
  nested dissection.''\hskip 1em plus 0.5em minus 0.4em\relax Georgia Institute
  of Technology, 2006.

\bibitem{geiger2012we}
A.~Geiger, P.~Lenz, and R.~Urtasun, ``Are we ready for autonomous driving? the
  kitti vision benchmark suite,'' in \emph{2012 IEEE Conference on Computer
  Vision and Pattern Recognition}.\hskip 1em plus 0.5em minus 0.4em\relax IEEE,
  2012, pp. 3354--3361.

\bibitem{nicholson2018quadricslam}
L.~Nicholson, M.~Milford, and N.~S{\"u}nderhauf, ``Quadricslam: Dual quadrics
  from object detections as landmarks in object-oriented slam,'' \emph{IEEE
  Robotics and Automation Letters}, vol.~4, no.~1, pp. 1--8, 2018.

\bibitem{liao2020object}
Z.~Liao, W.~Wang, X.~Qi, X.~Zhang, L.~Xue, J.~Jiao, and R.~Wei,
  ``Object-oriented slam using quadrics and symmetry properties for indoor
  environments,'' \emph{arXiv preprint arXiv:2004.05303}, 2020.

\bibitem{mousavian20173d}
A.~Mousavian, D.~Anguelov, J.~Flynn, and J.~Kosecka, ``3d bounding box
  estimation using deep learning and geometry,'' in \emph{Proceedings of the
  IEEE Conference on Computer Vision and Pattern Recognition}, 2017, pp.
  7074--7082.

\bibitem{yang2019cubeslam}
S.~Yang and S.~Scherer, ``Cubeslam: Monocular 3-d object slam,'' \emph{IEEE
  Transactions on Robotics}, vol.~35, no.~4, pp. 925--938, 2019.

\bibitem{hou20193d}
J.~Hou, A.~Dai, and M.~Nie{\ss}ner, ``3d-sis: 3d semantic instance segmentation
  of rgb-d scans,'' in \emph{Proceedings of the IEEE Conference on Computer
  Vision and Pattern Recognition}, 2019, pp. 4421--4430.

\bibitem{qi2019deep}
C.~R. Qi, O.~Litany, K.~He, and L.~J. Guibas, ``Deep hough voting for 3d object
  detection in point clouds,'' in \emph{Proceedings of the IEEE International
  Conference on Computer Vision}, 2019, pp. 9277--9286.

\bibitem{yang20203dssd}
Z.~Yang, Y.~Sun, S.~Liu, and J.~Jia, ``3dssd: Point-based 3d single stage
  object detector,'' in \emph{Proceedings of the IEEE/CVF Conference on
  Computer Vision and Pattern Recognition}, 2020, pp. 11\,040--11\,048.

\bibitem{hall2020probabilistic}
D.~Hall, F.~Dayoub, J.~Skinner, H.~Zhang, D.~Miller, P.~Corke, G.~Carneiro,
  A.~Angelova, and N.~S{\"u}nderhauf, ``Probabilistic object detection:
  Definition and evaluation,'' in \emph{The IEEE Winter Conference on
  Applications of Computer Vision}, 2020, pp. 1031--1040.

\bibitem{talbot2020benchbot}
B.~Talbot, D.~Hall, H.~Zhang, S.~R. Bista, R.~Smith, F.~Dayoub, and
  N.~S{\"u}nderhauf, ``Benchbot: Evaluating robotics research in photorealistic
  3d simulation and on real robots,'' \emph{arXiv preprint arXiv:2008.00635},
  2020.

\end{thebibliography}
\bibliographystyle{ieee}

\end{document}